\documentclass[11pt]{article}
\usepackage{geometry}
\usepackage{coling2020}
\usepackage{times}
\usepackage{url}
\usepackage{latexsym}
\usepackage{microtype}
\usepackage{graphicx}
\usepackage{amssymb}
\usepackage{array}
\usepackage{epstopdf}
\usepackage{subfig}

\usepackage{tikzsymbols}
\usepackage{subfig}
\colingfinalcopy 

\title{HPCC-YNU at SemEval-2020 Task 9: A Bilingual Vector Gating Mechanism for Sentiment Analysis of Code-Mixed Text}

\author{Jun Kong, Jin Wang \and Xuejie Zhang \\
  School of Information Science and Engineering \\
  Yunnan University \\
  Kunming, China \\
  {\tt Contact:\{wangjin, xjzhang\}@ynu.edu.cn} \\}

\date{}

\begin{document}
\maketitle
\begin{abstract}
  It is fairly common to use code-mixing on a social media platform to express opinions and emotions in multilingual societies. The purpose of this task is to detect the sentiment of code-mixed social media text. Code-mixed text poses a great challenge for the traditional NLP  system, which currently uses monolingual resources to deal with the problem of multilingual mixing. This task has been solved in the past using lexicon lookup in respective sentiment dictionaries and using a long short-term memory (LSTM) neural network for monolingual resources. In this paper, we (my codalab username is kongjun) present a system that uses a bilingual vector gating mechanism for bilingual resources to complete the task. The model consists of two main parts: the vector gating mechanism, which combines the character and word levels, and the attention mechanism, which extracts the important emotional parts of the text. The results show that the proposed system outperforms the baseline algorithm. We achieved fifth place in Spanglish and 19th place in Hinglish.The code of this paper is availabled at :
   \url{https://github.com/JunKong5/Semveal2020-task9}
\end{abstract}
\blfootnote{
    %
    %
    \hspace{-0.65cm}  
    %
    This work is licensed under a Creative Commons Attribution 4.0 International License. License details: http://creativecommons.org/licenses/by/4.0/.
    %
    %
    %
}

\section{Introduction}
\label{intro}

%
%

Sentiment analysis~\cite{WANG201893,wang-etal-2019-investigating} of social media data has been a hot research topic in the field of text in recent years. The emotion expressed in a phrase or sentence allows us to identify a person's point of view. Furthermore, the sentiment analysis of social media is critical to business and the government. With the integration of multiculturalism, there are many code-mixed texts on social media platforms. Code-mixing is a phenomenon in which two or more language units are mixed in one sentence, especially in multilingual societies around the world. Code-mixing specifically refers to the use of words, phrases, clauses and other language units in different languages at the sentence level. The purpose of Sentiment Analysis for Code-Mixed Social Media Text~\cite{patwa2020sentimix} is to analyze the sentiment of code-mixed text on social media platforms. The sentiment polarities of sentences include the following: positive, neutral and negative.

Compared to monolingual sentiment analysis~\cite{8930925}, coded-mixed text sentiment analysis is difficult due to the following reasons: (1) the language complexity of code-mixed content is exacerbated by spelling changes, slangs and non-compliance with formal grammar; (2) traditional semantic analysis methods cannot capture the meaning of code-mixed sentences; (3) most previous studies are focused on a single language, which ignores the phenomenon of code-mixing; (4) some words with the same spelling may have completely different meanings in different languages.

For this task, previous works have mostly focused on applying pre-trained word embedding on monolingual resources as the input features. Then, these features were be put into deep neural networks. Among the deep learning approaches, sub-word level representations in convolutional neural network (CNN)~\cite{Vieira2014} based on the long short-term memory (LSTM)~\cite{Hochreiter2016} architecture was presented by Joshi et al.~\shortcite{joshi-etal-2016-towards}. Others used features such as GloVe~\cite{Pennington2014} word embeddings with 300 dimensions. Furthermore, they trained an ensemble model that contains a linear support vector machine (SVM), logistic regression and random forest to detect the sentiments.

In this paper, we propose a vector gating mechanism to combine multiple monolingual word-level and char-level embeddings in a novel architecture. The char-BiLSTM layer is used to capture the character-level information and the word feature representation is generated by bilingual embedding. Then, the combined representation is processed using BiLSTM based Attention. In our model, the BiLSTM is used to capture the long-term dependencies between bilingual word sequences and character sequences. The gating mechanism can effectively combine character level and word level information, namely, the proposed model can precisely capture the emotional expression of code-mixed text. Our submission ranked fifth in Spanglish and 19th in Hinglish.

\begin{figure*}[!t]
\setlength{\belowcaptionskip}{-0.5cm}
\centering
\includegraphics[width=5.0in]{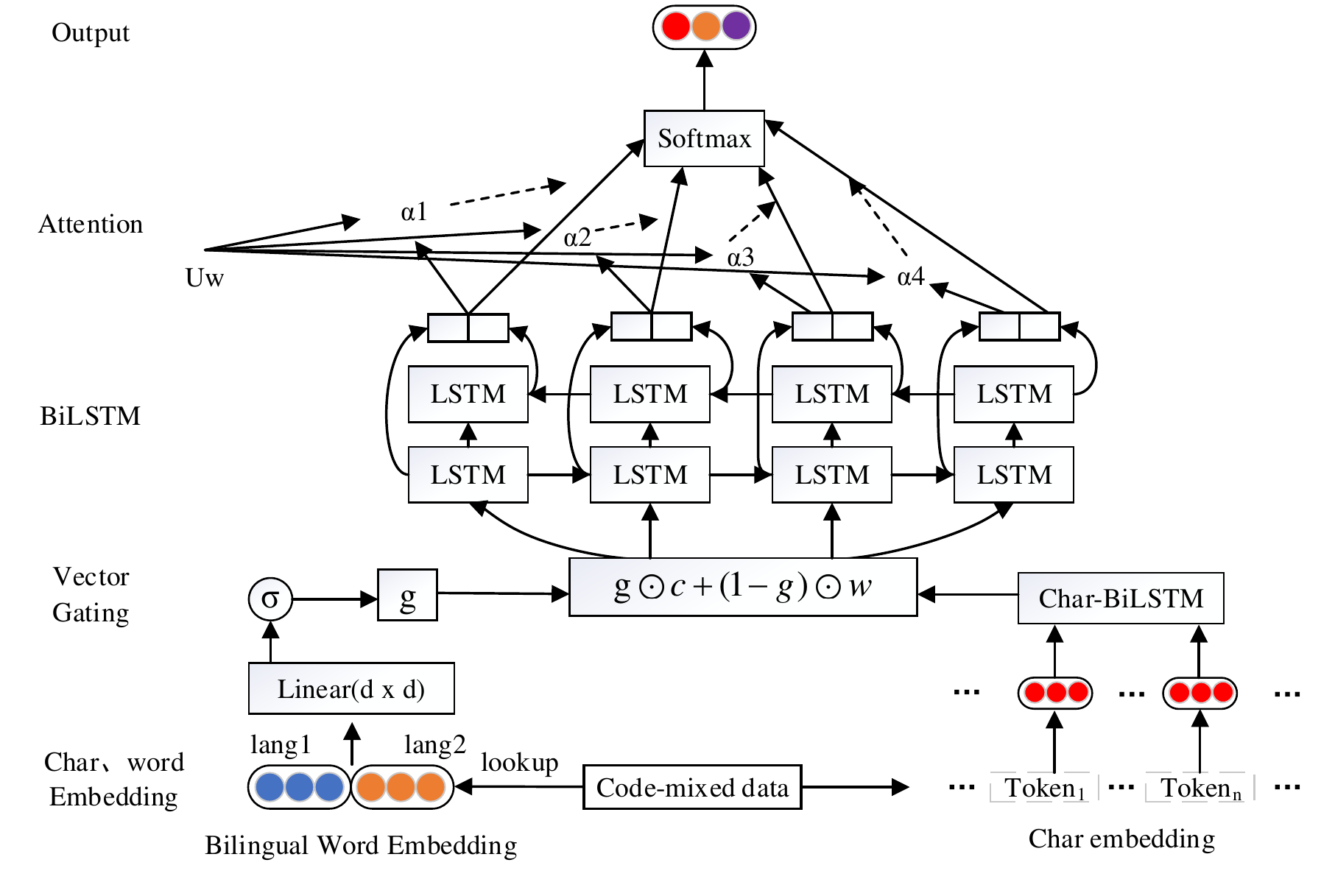}
\caption{Overall architecture of our model.}
\label{Figure2}
\end{figure*}

The rest of this paper is organized as follows. Section 2 describes the overall structure of our model and the gating mechanism. Then, the comparative experimental results are presented in section 3. Finally, the conclusions are finally drawn in section 4.

\section{Bilingual Vector Gating Model}

Figure 1 presents the overall architecture of our model. It uses preprocessed text as input to the model. In addition to taking word embeddings as the input, we also separate the words into characters and feed the character embeddings to the model. The word and character level representations are composed of bilingual pre-trained word vectors and the output of the char-BiLSTM layer. Then, the character- and word-level features are combined using the vector gating mechanism. Finally, a bi-directional LSTM (BiLSTM) with attention is used to calculate the gating vector to obtain the final result.

\subsection{Char-BiLSTM Embedding}

Character embedding is widely used in many NLP tasks. Character embedding can handle non-English words and misspelled words. Character embedding helps to improve the performance of word embedding in NLP tasks. At the char-level, each token is represented as a sequence of characters. Character embedding is initialized by using uniformly distributed random \textit{d}-dimensional vectors. BiLSTM is transformed from a bidirectional RNN~\cite{SchusterBidirectional}. The BiLSTM architecture is used to learn the character-based representation of each token. Figure 1 shows the model architecture. BiLSTM consists of both forward and backward LSTM, which capture the contextual relationships between the characters of each token. The LSTM consists of three gates, including the input gate $i_t$, the forget gate $f_t$ and the output gate $o_t$. The hidden state $h_t$ is calculated using the following equations:
\vspace{-0.2cm}
\begin{itemize}
\item Gates:
    \vspace{-0.1cm}
    \begin{equation}
    \begin{array}{l}
    {f_t} = \sigma ({W_f} \cdot \left[ {{h_{t - 1}},{x_t}} \right] + {b_f})\\
    {i_t} = \sigma ({W_i} \cdot \left[ {{h_{t - 1}},{x_t}} \right] + {b_i})\\
    {o_t} = \sigma ({W_o} \cdot \left[ {{h_{t - 1}},{x_t}} \right] + {b_o})
    \end{array}
    \label{eq:2}
    \end{equation}
    \vspace{-0.7cm}
\item Transformation:
    \vspace{-0.1cm}
    \begin{equation}
    \ {\tilde C_t} = \tanh ({W_c} \cdot \left[ {{h_{t - 1}},{x_t}} \right] + {b_c})
    \label{eq:3}
    \end{equation}
    \vspace{-1cm}
\item State update:
    \vspace{-0.1cm}
    \begin{equation}
    \begin{array}{l}
    {C_t} = {f_t} * {C_{t - 1}} + {i_t} * {{\tilde C}_t}\\
    {h_t} = {o_t} * \tanh ({C_t})
    \end{array}
    \label{eq:4}
    \end{equation}
    \vspace{-0.5cm}
\end{itemize}

\noindent where $x_t$ is the input vector, $\sigma$ denotes the sigmoid function, $W$ and $b$ are cell parameters. $\tilde C_t$ represents the candidate values that are created by a tanh activation function and finally update the cell state $C_t$. Each character is embedded in a \textit{d}-dimensional vector. We then use it as the input to BiLSTM to get a representation of each token. The output of BiLSTM is a concatenation of both the forward hidden state \overrightarrow {{h_t}} and backward hidden state \overleftarrow {{h_t}}, which is defined as,

\vspace{-0.3cm}
\begin{equation}
y = \overrightarrow {{h_t}}  \otimes \overleftarrow {{h_t}}
\label{eq:5}
\end{equation}
\vspace{-0.8cm}

\subsection{Bilingual Word Embedding}

On the word-level, word embedding is used to represent each token. These embeddings are bilingual continuous low-dimensional vectors. To create the shared vocabulary, the English and other word vectors (Spanish or Hindi) are concatenated. We use GloVe~\cite{Pennington2014} that is a 300-dimensional and pre-trained English word vector, and use the Spanish and Hindi FastText 300-dimensions word vectors~\cite{grave-etal-2018-learning} that are trained using Common Crawl and Wikipedia. Each code-mixed text finds the corresponding word vector from the shared dictionary for each token.

\subsection{Vector Gating Mechanism}

Since word-level embeddings do not account for out-of-vocabulary words in code-mixed texts, we combine character-level and word-level word representations together to get better results than using only char-level representations. The vector gating mechanism is used to connect character and word-level representations as in~\cite{balazs-matsuo-2019-gating}. As illustrated in Figure 1, the vector gating mechanism learns how to independently weight the dimensions of each vector at a fine-grained level. Furthermore, the vector gating mechanism differs from the traditional scalar gating mechanism and works on each dimension of the character and word vectors. The vector gating mechanism is expressed as follows:

\vspace{-0.2cm}
\begin{equation}
\begin{array}{l}
{g_i} = \sigma (Wv_i^{(w)} + b)\\
{v_i} = {g_i} \odot v_i^{(c)} + (1 - {g_i}) \odot v_i^{(w)}
\end{array}
\label{eq:6}
\end{equation}
\vspace{-0.1cm}

\noindent where each token can be represented as a vector $
v_{i}^{(w)} \in \mathbb{R}^{d} $ by Bilingual pre-trained word vectors. The vector $v_{i}^{(c)} \in \mathbb{R}^{d}$ is built from the characters of each token. $W \in \mathbb{R}^{d \times d}$ and $b \in \mathbb{R}^{d}$ are trainable parameters, ${g_i} \in {(0,1)^d}$ , $\sigma$ is the element-wise sigmoid function, $\odot$ is the element-wise product for vectors, and $1 \in \mathbb{R}{^d}$ is a vector of ones.

\subsection{BiLSTM based Attention}

The output vector of the vector gating mechanism is then fed into the bidirectional LSTM structure. The essence of the attention mechanism is modeled after human visual attention. An attention mechanism~\cite{NIPS2017_7181} is used in most NLP tasks. It assigns different weights to each token of the code-mixed text so that important contextual information can be captured. We combined an attention mechanism~\cite{Wang2016} with BiLSTM as represented by Figure 1. The attention mechanism is expressed as follows:

\vspace{-0.2cm}
\begin{equation}
\begin{array}{l}
{e_{_t}} = \tanh ({W_e}{h_t} + {b_e})\\
{\alpha _t} = softmax({e_{_t}})
\end{array}
\label{eq:7}
\end{equation}
\vspace{-0.2cm}

\noindent where $h_t$ represents the hidden state of the BiLSTM output layer, $W_e$ and $b_e$ are trainable parameters, and $\alpha _t$ represents the weight of each input. The remaining steps are consistent with the BiLSTM.

\section{Experimental Results}
{\bf Datasets}. The organizers collected and annotated a dataset from social media platforms such as Twitter and Facebook (Patwa et al., 2020). The datasets consist of two parts: Hinglish and Spanglish code-mixed texts. Table 1 shows the details of the Spanglish and Hinglish datasets. These tweets are tagged based on their word-level language, emoticons, special characters, etc. Furthermore, the sentiment polarity of each tweet was classified as negative, neutral or positive. Examples of code-mixed texts are shown in the Table 2.

\begin{table}[!t]
\begin{center}
\renewcommand{\thefootnote}{\fnsymbol{footnote}}
\begin{tabular}{|c|c|c|c|}
\hline
{\bf Datasets} & {\bf Train } & {\bf Dev} & {\bf Test} \\
\hline
Hinglish & 14000 & 3000 & 3000 \\
 Spanlish & 12002 & 2998 & 3789 \\
\hline
\end{tabular}
\end{center}
\caption{Data Statistics for the Hinglish and Spanglish Datasets. }
\end{table}

\begin{table}
\centering
\small
\renewcommand{\thefootnote}{\fnsymbol{footnote}}
\begin{tabular}{|l|l|}
\hline
\multicolumn{1}{|c|}{\bf Code-mixed text} & \multicolumn{1}{c|}{\bf Label } \\
\hline
Hinglish: &\\I'm$_{Eng}$  poor$_{Eng}$  poor-ever$_{Eng}$  happy$_{Eng}$   kase$_{Hin}$  God$_{Hin}$  is$_{Eng}$   with$_{Eng}$   me$_{Hin}$  
\includegraphics[width=4mm]{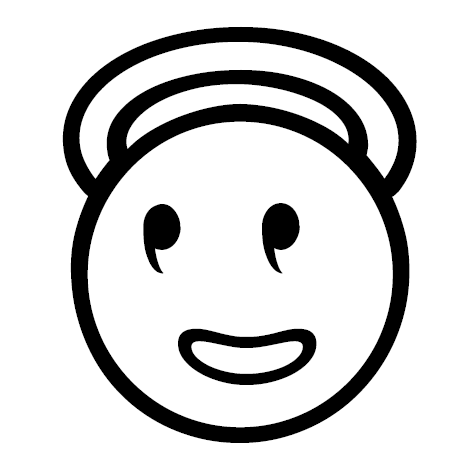}$_{EMT}$
&   negative \\
\hline
Spanglish: &\\
$\rm @\_2Gay4You_{\_other}$   ha$_{lang2}$   pos$_{ambiguous}$  have$_{lang1}$  fun$_{lang1}$   its$_{lang1}$   pretty$_{lang1}$   te$_{lang2}$ &\\
subes$_{lang2}$ al$_{lang2}$  horse$_{lang1}$  its$_{lang1}$  cute$_{lang1}$  lol$_{lang1}$
& positive \\
\hline
\end{tabular}
\caption{Examples of Hinglish and Spanglish code-mixed texts.}
\end{table}

\begin{table}[!t]
\begin{center}
\renewcommand{\thefootnote}{\fnsymbol{footnote}}
\begin{tabular}{|c|c|c|}
\hline
{\bf Model} & {\bf Hinglish(\textit{F}1-score) } & {\bf Spanglish (\textit{F}1-score)} \\
\hline
Char-embedding & 0.492 &0.678 \\
\hline
Sub-word & 0.564 & 0.682  \\
\hline
Vector Gating & 0.681 & 0.725 \\
\hline
Bilingual Vector Gating  & 0.690  & 0.753 \\
\hline
\end{tabular}
\end{center}
\caption{Hinglish and Spanglish model results. }
\end{table}

\begin{figure*}[!t]
\setlength{\belowcaptionskip}{-0.3cm}
\centering

\subfloat[Performance of different numbers of epochs]{          \label{(a)the performance of different epoch}
         \begin{minipage}[c]{.5\linewidth}
             \centering
             \includegraphics[width=8cm]{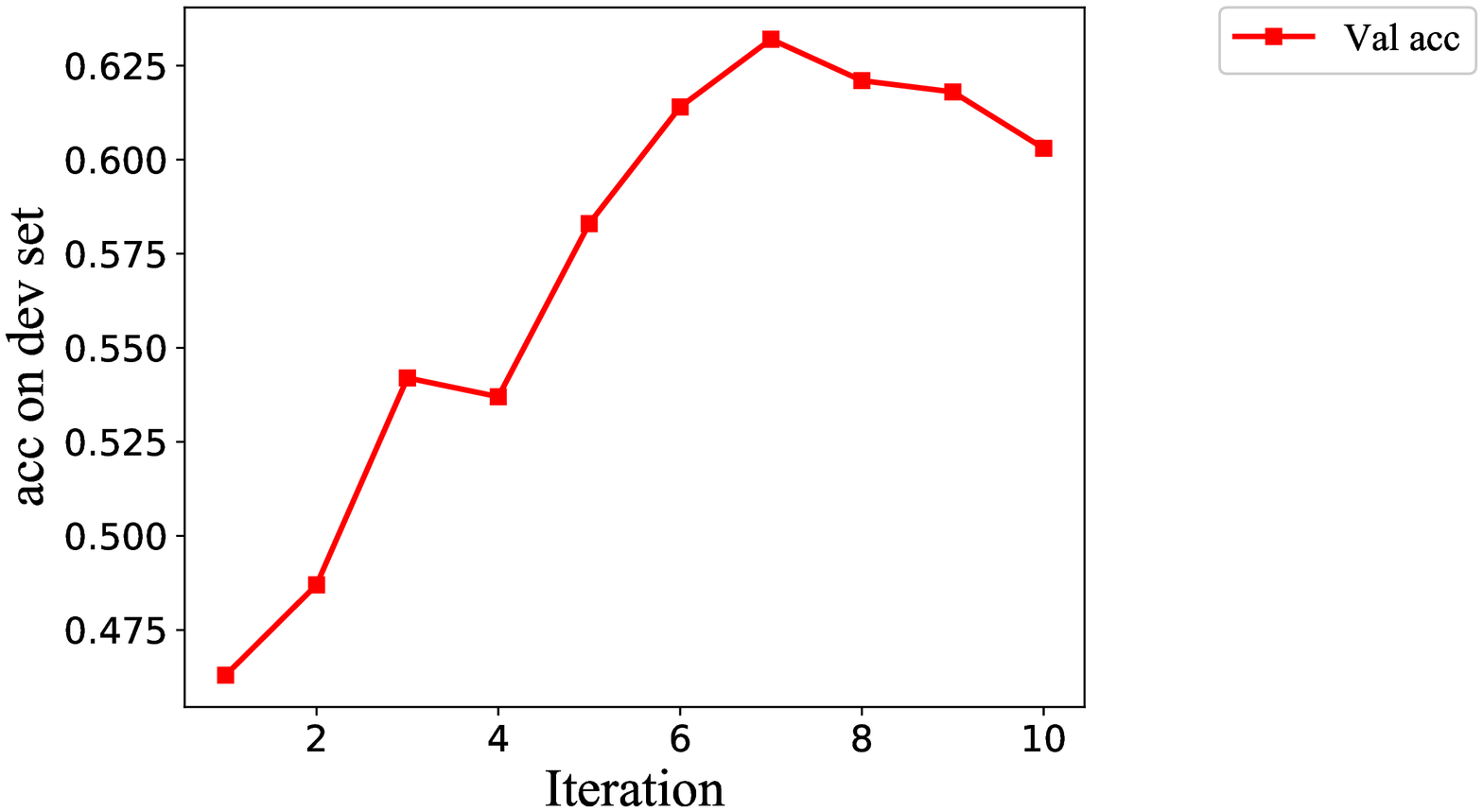}
          \end{minipage}
    }
\subfloat[Performance of different batch sizes]{          \label{(b)}
         \begin{minipage}[c]{.5\linewidth}
             \centering
             \includegraphics[width=8cm]{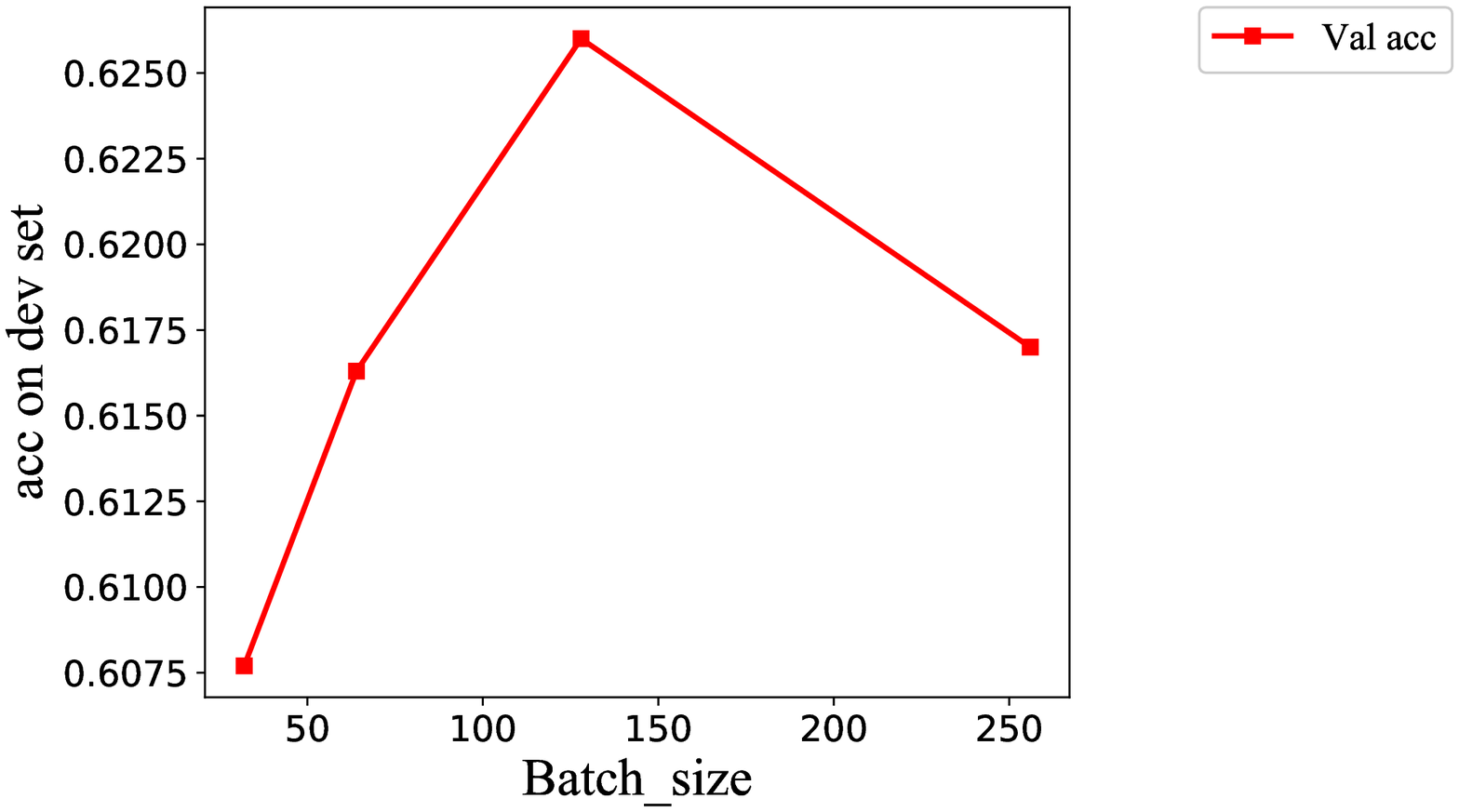}
          \end{minipage}
    }

\caption{Epoch and batch size parameter selection for the dev set.}
\label{Figure1}
\end{figure*}

\vspace{0.2cm}
\noindent {\bf Evaluation Metrics}. The system is evaluated by calculating the average $F_{1}$-score across the positive, negative, and neutral expressions. The final ranking would be based on the average $F_{1}$-score. The average $F_{1}$-score ranges from1 to 0 and is defined as, 

\vspace{-0.3cm}
\begin{equation}
{F_1} - score = 2 \times \frac{{P * R}}{{(P + R)}}
\label{eq:8}
\end{equation}
\vspace{-0.3cm}

\noindent where \textit{P} denotes the precision and \textit{R} denotes the recall. A higher $F_{1}$-score indicates better model prediction performance.

\vspace{0.2cm}
\noindent {\bf Implementation Details}. Twitter data are informal social media texts that always contain many noisy features. Effective preprocessing can reduce the number of OOV words and improve the performance of the model. Therefore, the texts were preprocessed using the following procedures before model training: 

$\bullet$  All URLs were removed, @someones are replaced with \textit{user}, and \#somethings are replaced with \textit{hashtag}. 

$\bullet$  All uppercase letters were converted to lowercase letters.

$\bullet$  Strings of repetitive of marks (. ? !) and contractions were replaced by their equivalents. 

$\bullet$  A dictionary of slang terms and their equivalents was created and used to replaces the slangs. 

$\bullet$  Emoticons in text were replaced by their corresponding meaning. For example, ‘\includegraphics[width=4mm]{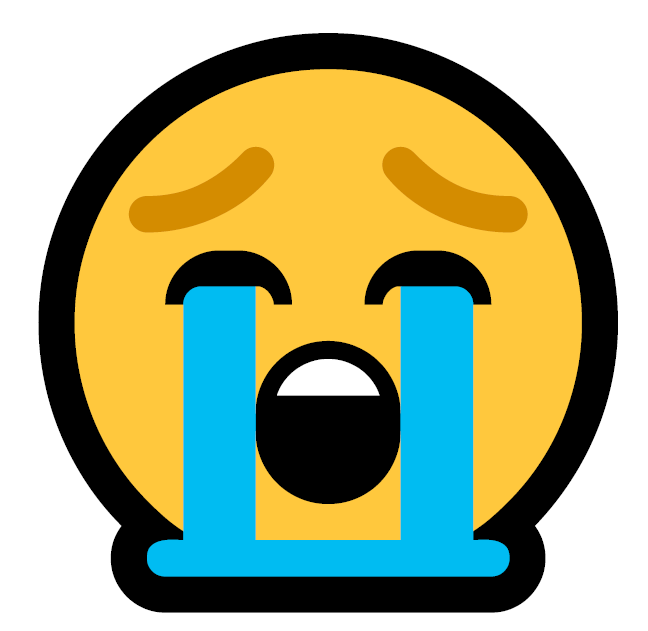}’ was replaced by ‘\textit{Loudly Crying Face}’. 

The experiments were conducted on Keras with a TensorFlow backend. The two different language pre-trained word vectors and the char-embedding are used to train our model.

\vspace{0.2cm}
\noindent {\bf Parameters Fine-tuning}. The parameters are tuned on the training and development sets. Early stopping is used to determine the number of iterations during training. If the loss is not improved within 3 epochs, the training process will be terminated. The number of epochs and batch size affect the final performance of our proposed model. In the training steps, the performance on the dev set with different numbers of iterations is shown in Figure 2(a). When the number of epochs exceeded 7, the performance of the model began to decline. This is probably caused by overfitting. This demonstrates that setting the number of epochs to 7 can bring the best performance. The batch size is set to 128 because the performance of model was improved on the dev sets, as shown in figure 2(b). We also set the dropout rate to 0.25 to prevent overfitting. The optimizer was Adamax with the categorical cross-entropy as the loss function. The char-embedding size is also set to 150. The dimension of the hidden layer in LSTM is 150. 

\vspace{0.2cm}
\noindent {\bf Comparative Results}.After submitting the results three times, the following experiments were additionally done. The comparative experimental results of Hinglish and Spanglish are shown in Table 3. As indicated, the experimental results of our proposed bilingual vector gating model achieved the highest scores in both Spanglish and Hinglish. In the Spanglish dataset, the $F_{1}$-score of the proposed Bilingual Vector Gating model is 3\% higher than that of the Vector Gating model. Additionally, the proposed model achieved 69.0\% in Hinglish, which is 13.6\% higher than that of the sub-word model and 19.8\% higher than that of the char-embedding model.

\vspace{0.2cm}
\noindent {\bf Discussion}. As illustrated in Table 3, the bilingual Vector Gating model performs better than the Vector Gating model. That is because of the use of bilingual resources. The Vector Gating model performs better than Sub-word. Therefore, the results show the effectiveness of the vector gating mechanism that combines the word-level and char-level embeddings. From equation (5), \textbf{g} = \textbf{0} means only the word-level, and \textbf{g} = \textbf{1} means only the character-level. The character and word levels are well integrated by \textbf{g} (the vector gate). The vector gating mechanism precisely captures the emotional expression of code-mixed text.

\section{Conclusions}

In this paper, we describe the system we submitted to the SemEval-2020 task 9 on Sentiment Analysis for Code-Mixed Social Media Text. The proposed model that uses the vector gating mechanism combines the bilingual word vector and char-embedding to solve the challenge of code-mixed text. Our introduced model achieved good performance according to the experimental results. In future work, we will attempt to introduce BERT to draw more useful sentiment information.

\section*{Acknowledgements}
This work was supported by the National Natural Science Foundation of China (NSFC) under Grant No. 61966038, 61702443 and 61762091. The authors would like to thank the anonymous reviewers for their constructive comments

\bibliographystyle{coling}
\bibliography{semeval2020}

\end{document}